\documentclass[letterpaper,10pt,conference]{template/ieeeconf}
\newif\ifanonymousmode
\anonymousmodefalse 

\newcommand{\realauthorname}{Jiajie Zhang, Yankai Xiang, Changhao Chen}

\ifanonymousmode

  \newcommand{\paperauthorblock}{\anonauthorname}
  
  \newcommand{\paperpdfauthor}{\anonauthorname}
  \newcommand{\paperpdfsubject}{Anonymous ICRA 2027 submission}

\else

  \IEEEoverridecommandlockouts

  \newcommand{\paperauthorblock}{%
    Jiajie Zhang$^{1,\dagger}$, Yankai Xiang$^{2,\dagger}$, and Changhao Chen$^{1,*}$%
    \thanks{\raggedright $^{1}$Jiajie Zhang and Changhao Chen are with the Hong Kong
      University of Science and Technology (Guangzhou), Guangzhou, China
      {\tt\small jiajiezhang@hkust-gz.edu.cn},
      {\tt\small changhaochen@hkust-gz.edu.cn}}%
    \thanks{\raggedright $^{2}$Yankai Xiang is with ShanghaiTech University, Shanghai,
      China {\tt\small xiangyk2025@shanghaitech.edu.cn}}%
    \thanks{$^{\dagger}$These authors contributed equally to this work.}%
    \thanks{$^{*}$Corresponding author.}%
  }
  
  \newcommand{\paperpdfauthor}{\realauthorname}
  \newcommand{\paperpdfsubject}{ICRA 2027 submission draft, author-identifying version}

\fi
\usepackage{amsmath,amssymb}
\usepackage{booktabs}
\usepackage{graphicx}
\usepackage{array}
\usepackage{tabularx}
\usepackage{xcolor}
\usepackage{microtype}
\usepackage{cite}

\makeatletter
\renewcommand{\thetable}{\@arabic\c@table}
\renewcommand{\fnum@table}{\textit{Table~\thetable}}
\long\def\@makecaption#1#2{%
  \ifx\@captype\@IEEEtablestring%
    \setbox\@tempboxa\hbox{\footnotesize\normalfont #1.~#2}%
    \ifdim \wd\@tempboxa >\hsize%
      \parbox[t]{\hsize}{\footnotesize\normalfont\noindent #1.~#2}%
    \else%
      \hbox to\hsize{\hfil\box\@tempboxa\hfil}%
    \fi%
    \@IEEEtablecaptionsepspace%
  \else
    \@IEEEfigurecaptionsepspace%
    \setbox\@tempboxa\hbox{\footnotesize #1.~~ #2}%
    \ifdim \wd\@tempboxa >\hsize%
      \setbox\@tempboxa\hbox{\footnotesize #1.~~ }%
      \parbox[t]{\hsize}{\footnotesize \noindent\unhbox\@tempboxa#2}%
    \else%
      \ifcenterfigcaptions \hbox to\hsize{\footnotesize\hfil\box\@tempboxa\hfil}%
      \else \hbox to\hsize{\footnotesize\box\@tempboxa\hfil}%
    \fi\fi\fi}
\makeatother
\usepackage{url}
\usepackage[hidelinks]{hyperref}
\usepackage{flushend}

\definecolor{cmablue}{HTML}{0072B2}
\definecolor{cmagreen}{HTML}{009E73}
\definecolor{cmaorange}{HTML}{E69F00}
\definecolor{cmared}{HTML}{D55E00}
\definecolor{cmagray}{HTML}{70757A}

\graphicspath{{figures/}}

\newcommand{\method}{CMA}

\newcommand{\ci}[2]{\ensuremath{[#1,#2]}}

\hypersetup{
  pdfauthor={\paperpdfauthor},
  pdftitle={Memory That Changes Action Is Not Memory That Guides It: Counterfactual Auditing of History-Conditioned Robot Policies},
  pdfsubject={\paperpdfsubject},
  pdfkeywords={robot memory, counterfactual evaluation, partial observability, reliability, causal intervention}
}

\ifdefined\pdfinfoomitdate
\fi
\ifdefined\pdftrailerid
  \pdftrailerid{}
\fi

\makeatletter
\def\bstctlcite{\@ifnextchar[{\@bstctlcite}{\@bstctlcite[@auxout]}}
\def\@bstctlcite[#1]#2{\@bsphack
  \@for\@citeb:=#2\do{%
    \edef\@citeb{\expandafter\@firstofone\@citeb}%
    \if@filesw\immediate\write\csname #1\endcsname{\string\citation{\@citeb}}\fi}%
  \@esphack}
\makeatother

\title{Memory That Changes Action Is Not Memory That Guides It:\\
Counterfactual Auditing of History-Conditioned Robot Policies
}

\author{\paperauthorblock}

\begin{document}
\bstctlcite{IEEEexample:BSTcontrol}

\maketitle
\thispagestyle{empty}
\pagestyle{empty}

\begin{abstract}
\looseness=-1
A robot returning a block to its origin tray may encounter two task-consistent pasts that reconverge to the same current input but warrant different actions. Yet memory-policy evaluations often rely on task success or action change under memory perturbation, neither of which establishes that memory guides the decision. We propose the \textbf{Counterfactual Memory Audit (CMA)}, an evaluation protocol that crosses two histories at a verified-identical present, queries a frozen policy under common randomness, and evaluates each saved action under both pasts. This separates memory sensitivity, warranted choice, matched-world physical value, and per-pair reliability. On Mem-0, every audited Put Back pair changes action, but only $20/64$ pairs are fully reliable; at a later Swap decision, all paired actions change while both memories select the same branch. Native interventions further show closed-loop influence: replacing the history bank redirects behavior toward the replaced content, while restoring a 4096-byte protected anchor recovers $38.9$ points of Swap success lost to injected bank faults. On a dual-arm physical platform, memory changes saved actions, yet five of nine completed Put Back manipulations reach the wrong target. These results show that a robot can remember and react without reliably using memory to choose the behavior its past warrants. CMA provides a decision-level audit for distinguishing these cases.
\end{abstract}

\section{Introduction}
\label{sec:intro}

Robot manipulation is acquiring memory. Recent robot learning policies retain information in many forms, from recurrent states to retrieval banks and episodic stores
\cite{li2026rememvla,cherepanov2026muvla,hu2025pam,sridhar2026memer,torne2026mem,li2026bridgevlapp},
while a growing set of benchmarks evaluates them under delayed evidence, long horizons, and memory interference
\cite{fang2025sam2act,cherepanov2025mikasa,chen2026rmbench,dai2026robomme,lei2026robomemarena}.
Yet these evaluations ultimately reduce memory use to task-level outcomes. This invites a tempting inference: if adding memory improves the score, the policy must be using that memory. But such evidence never examines whether memory determines an individual decision.

\begin{figure}[t]
  \centering
  \includegraphics[width=\columnwidth]{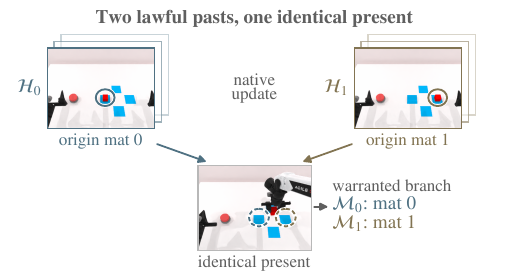}
  \caption{\textbf{The audited decision: two lawful pasts, one identical present.} Frames from one frozen Put Back pair (illustrative). Each task-consistent history ($\mathcal{H}_0$, $\mathcal{H}_1$) updates its native memory ($\mathcal{M}_0$, $\mathcal{M}_1$) before reconverging to the same present. Stacked outlines denote the stored multi-frame executed history. Solid rings mark each history's origin mat; dashed rings mark the same mats at the present, both empty, making the warranted branch invisible from the current observation alone.}
  \label{fig:premise}
\end{figure}

\looseness=-1
Consider what a task-level score cannot reveal. A robot picks up a block from one of two trays and must later return it to its \emph{original} tray. At the return decision, two lawful pasts, generated entirely by the task's own motion primitives, can reconverge to the same image, proprioception, and instruction, with both candidate trays now empty (Fig.~\ref{fig:premise}). The correct branch is therefore absent from the current input; only memory distinguishes the pasts that warrant different actions. This is a concrete instance of perceptual aliasing in a partially observable control problem
\cite{kaelbling1998planning}, precisely the setting in which memory should matter. A frozen policy may nevertheless complete the task reliably. The central question remains unanswered: \emph{at this particular decision, did the remembered content cause the policy to choose the branch it warranted?}

Existing evaluation protocols answer related but narrower questions. Episode success measures overall competence, yet a policy may recover from an incorrect local decision through subsequent replanning; success therefore certifies the outcome, not the decision. Retrieval and probing accuracy establish that historical information is accessible or decodable, but accessibility does not imply task-relevant use
\cite{ravichander2021probing}. Memory ablations reveal aggregate dependence across episodes, but do not identify the direction of an individual decision. Action distance detects behavioral change, but an action may change without changing the underlying task branch. None of these metrics directly establishes whether memory caused the warranted decision at a fixed present.


\begin{figure*}[t]
  \centering
  \includegraphics[width=0.95\textwidth]{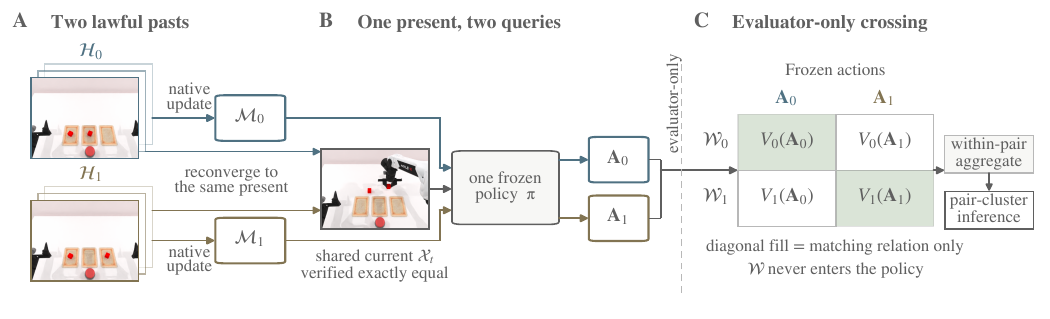}
    \caption{\textbf{CMA crosses lawful memories at one fixed present.}
  (A) Two lawful histories from one illustrative later-Swap pair differ only in which mats
  hold a block and update two native memories (stacked outlines: stored history). (B) Both
  queries receive the verified-exactly-equal shared current $\mathcal{X}_t$; the frozen policy is
  queried once per memory under common randomness. (C) The saved actions cross the
  evaluator-only boundary (dashed) and are scored in both worlds, which never enter the
  policy; the four scores aggregate within pair before pair-cluster inference.}
  \label{fig:method}
\end{figure*}

To address this gap, we propose the \textbf{Counterfactual Memory Audit (\method{})}, an evaluation protocol that makes this
question measurable for a frozen robot policy (Fig.~\ref{fig:method}). We construct two task-consistent histories, update their respective native memories, and then reconverge to a verified-identical current input. The policy is queried once per memory under common randomness, and each resulting action is evaluated in both evaluator-only worlds, which are never exposed to the policy. This design yields a four-level audit, with each level asking a stricter question. First, \emph{sensitivity}: does changing memory change the action? Second, \emph{choice}: does the change favor the branch encoded by the memory? Third, \emph{matched-world value}: when memory matches the underlying world, does the selected action improve continuous physical value? Finally, \emph{reliability}: does this effect hold across every pair, both crossing directions, and every registered action seed? These are distinct verdicts: sensitivity does not imply correct choice, correct choice does not imply physical value, and average value does not imply reliable memory use.

\looseness=-1
Applied to frozen checkpoints, the audit dissociates the two standard behavioral signals: one task is correct on average yet unreliable pair by pair, while another changes every action without ever changing the selected branch. A dual-arm physical platform reproduces the same dissociation. At the same time, memory can exert closed-loop influence: replacing the history bank redirects every continuation toward the replaced content, while restoring a single 4096-byte protected component recovers 38.9 points of success lost to injected bank faults. These results motivate evaluating memory beyond episode-level outcomes and suggest a design principle for future systems: when memory is implemented as a declared, protectable component, its decision-level
influence can be audited, and its critical content selectively restored after faults.

Our contributions are:
\begin{itemize}
    \item We propose the Counterfactual Memory Audit (CMA), an evaluation protocol for history-conditioned manipulation policies that crosses two task-consistent histories at a verified-identical present and evaluates the resulting decision with four decision-level metrics absent from episode-level benchmark scores.
    \item We show that task success and action change alone are insufficient evidence of memory use, while audited memory can exert closed-loop influence: replacing memory redirects the robot toward the replaced content, and restoring a small protected component recovers success lost under injected memory-bank faults.
    \item We apply the audit to two separately analyzed frozen manipulation policies across tasks, decision phases, and four intervention classes, and reproduce the same crossing protocol on a dual-arm physical platform with measured resets.
\end{itemize}

\section{Related Work}
\label{sec:related}

\subsection{Memory Policies and Benchmarks}
Manipulation policies increasingly carry explicit memory, and its form varies widely:
recurrent states and recurrent queries that summarize history inside the network
\cite{li2026rememvla,cherepanov2026muvla}, adaptive working-memory recoding against
state ambiguity \cite{hu2025pam}, retrieval over stored experience
\cite{sridhar2026memer}, and memory-augmented VLA frameworks with episodic or
multi-scale stores \cite{torne2026mem,li2026bridgevlapp}. A parallel line of benchmarks asks whether such memory helps: MemoryBench scores spatial recall \cite{fang2025sam2act}, MIKASA organizes memory-intensive RL tasks \cite{cherepanov2025mikasa}, and RMBench, RoboMME, and RoboMemArena score manipulation policies episode by episode under delayed evidence, long horizons, and memory interference \cite{chen2026rmbench,dai2026robomme,lei2026robomemarena}. All of
these certify memory at the episode level: a policy is credited when it finishes tasks
designed to require history. Neither line identifies whether remembered content
governed one action at a fixed present, and both supply CMA's subjects: the frozen
checkpoints (Mem-0 from RMBench \cite{chen2026rmbench}, BridgeVLA++
\cite{li2026bridgevlapp}) and the tasks under audit.

\subsection{Counterfactual Tests of Memory Use}
A small family of evaluations intervenes directly on what a policy has stored. LIBERO-CF
tests a complementary counterfactual generalization question \cite{fang2026liberocf};
TRACE perturbs the executed routing path and reads out route and branch consistency
\cite{li2026trace}; TFP replaces hidden memory at a fixed current input but stops at
action variation, the first rung of the audit \cite{liang2026tfp}. The closest work,
Chameleon, keeps, swaps, or removes stored traces and reports counterfactual choice
accuracy, covering the audit's first two questions \cite{guo2026chameleon}. None
replaces complete lawful native histories, verifies an exactly equal current packet, and
scores the same saved actions under both evaluator worlds; CMA adds crossed physical
value and per-pair bidirectional reliability, closing two real gaps: a discrete choice
can hide different world value, and a positive average can hide pairs failing one
direction or one seed.

\subsection{Causal Attribution and Memory Intervention}

\looseness=-1
Causal misidentification and shortcut copying show that history can mislead as well as
inform \cite{dehaan2019causal,wen2020copycat,gao2026gmp}; those works motivate
training-time remedies, whereas CMA diagnoses a frozen policy. Probing separates
decodability from use \cite{ravichander2021probing}, and interchange interventions and
activation patching ask related causal questions at the representation level
\cite{geiger2021causal,zhang2024patching}; in frozen VLAs they find stored history
causally deployed mainly under degraded current frames \cite{liao2026present}, whereas
CMA audits the complementary regime, an intact present that cannot supply the branch,
and reads out closed-loop behavior rather than internal representation. Repeated-measure unit errors motivate CMA's pair-cluster inference and common randomness
\cite{agarwal2021statistical}. Online detection and scheduling around memory
\cite{zeng2026helm,chen2026aegis,gao2026agm} and counterfactual attribution with
rollback repair for text-agent memories \cite{tan2026memaudit,yu2026rollback} build
deployment mechanisms; we train no detector and test no adaptive timing, instead
restoring one declared tensor of a robot policy's memory carrier at a fixed boundary and
reading out physical task success.

\section{Counterfactual Memory Audit}
\label{sec:method}

\subsection{Audit Protocol}

\looseness=-1
Our proposed CMA protocol holds fixed everything the policy can see and changes only what it remembers. Throughout, light italic denotes scalars ($S$, $F$, $U$, $L$, and all indices), bold capitals denote matrices, and calligraphic capitals denote structured objects that are neither: histories, memory states, input packets, and evaluator worlds. Two task-consistent histories, $\mathcal{H}_0$ and $\mathcal{H}_1$, are generated entirely by the task's motion primitives and warrant semantic branches $0$ and $1$, respectively. Their native updates produce memory states $\mathcal{M}_0$ and $\mathcal{M}_1$, after which both histories reconverge to a verified-identical current packet $\mathcal{X}_t$ at the audited present $t$. The packet contains every non-memory input available to the policy, including raw and encoded images, proprioception, instruction, action tail, controller state, and clocks. The underlying physical state is controlled and verified to be equal across the two queries but is never exposed to the policy. We keep three entities separate by construction: a memory state $\mathcal{M}$ contains only the native history carrier; an action chunk $\mathbf{A}\in\mathbb{R}^{K\times d_a}$ stacks the $K$ serialized $d_a$-dimensional actions returned by one query, a single-action query being the case $K=1$; and the evaluator world $\mathcal{W}_w$ contains only branch-$w$ oracle and value semantics.

\looseness=-1
Pair eligibility is outcome-blind. It may depend on the generator, task legality, successful memory update, exact restoration, and current-state equality, but never on action sensitivity, decision direction, physical value, or task success. A failed equality check excludes the pair, and excluded pairs are never replaced after the action has been observed. Indexing eligible pair clusters by $p$ and registered action-seed repeats by $r$, the two queries of a pair share common action randomness through the scalar seed $\xi_r$, which fixes the policy's internal draw for repeat $r$. The frozen policy $\pi$ maps one current packet and one memory state to an action chunk, and is queried once per memory:
\begin{equation}
  \mathbf{A}_{pmr}=\pi\!\left(\mathcal{X}_{t,p},\,\mathcal{M}_{pm};\,\xi_r\right),
  \qquad m\in\{0,1\},
  \label{eq:action}
\end{equation}
where $\mathcal{X}_{t,p}$ is the shared current packet of pair $p$ at its audited present and $\mathcal{M}_{pm}$ is the memory produced by history $\mathcal{H}_{pm}$.

Each action is saved before any evaluator-world selection and then evaluated in both $\mathcal{W}_0$ and $\mathcal{W}_1$, which change only the oracle and value functional. Evaluator labels enter neither the policy query nor pair eligibility. The resulting crossing isolates the local role of memory at the audited present; it does not estimate how frequently such perceptual aliasing occurs in natural episodes.

\subsection{Decision-Level Audit Metrics}

Across pairs $p$ and action-seed repeats $r$, we first measure whether memory changes the action (i.e., action sensitivity):
\begin{equation}
 S=\mathbb{E}_{p,r}\,d\!\left(\mathbf{A}_{p0r},\mathbf{A}_{p1r}\right),
 \label{eq:sensitivity}
\end{equation}
where $\mathbb{E}_{p,r}$ is the empirical average over eligible pair clusters and their registered seeds, and $d$ is a preregistered executed-action distance. In the primary audit it is the RMSE over the executed chunk, $d(\mathbf{A}, \mathbf{A}')=\lVert\mathbf{A}-\mathbf{A}'\rVert_{\mathrm{F}}/\sqrt{K d_a}$
, and we write $S_{\mathrm{exec}}$ for $S$ under this distance. Sensitivity is unsigned: two actions may differ while approaching the same tray, choosing neither branch, or following the same branch.

A preregistered oracle is a scalar-valued map $Y_w$ on action chunks, $Y_w(\mathbf{A})\in\{-1,0,+1\}$: $+1$ when $\mathbf{A}$ selects the branch warranted in $\mathcal{W}_w$, $-1$ for the alternative, and $0$ for \emph{ambiguous}, a third outcome distinct from both error and success. In the primary Put Back audit, the oracle normalizes each target's signed block progress over the executed chunk by the inter-target separation and selects a branch only when one progress exceeds the other by the preregistered margin $\delta=0.05$. Other cohorts use their registered normalization and margin. Signed faithfulness scores each memory only in the world it matches,
\begin{equation}
 F=\mathbb{E}_{p,r}\tfrac12\!\left[Y_0(\mathbf{A}_{p0r})+Y_1(\mathbf{A}_{p1r})\right].
 \label{eq:faithfulness}
\end{equation}
We additionally report the raw correct-choice rate $p_{\mathrm{correct}}$ and the ambiguity rate $p_{\mathrm{ambiguous}}$, the fractions of matched-world decisions that the oracle scores $+1$ and $0$, pooled over both crossing directions, pairs, and seeds. Because $F=2p_{\mathrm{correct}}+p_{\mathrm{ambiguous}}-1$, $F$ is a compact signed summary rather than an additional source of information beyond the underlying outcome rates.

To quantify continuous physical benefit, let $V_w(\mathbf{A})$ denote the preregistered scalar normalized value of action chunk $\mathbf{A}$ in evaluator world $\mathcal{W}_w$, so that $\mathcal{W}_w=(Y_w,V_w)$ is the pair of branch-$w$ functionals. We define matched-world value as the World $\times$ Memory contrast
\begin{equation}
 U=\mathbb{E}_{p,r}\tfrac12[
 V_0(\mathbf{A}_{p0r})-V_0(\mathbf{A}_{p1r})+V_1(\mathbf{A}_{p1r})-V_1(\mathbf{A}_{p0r})].
 \label{eq:utility}
\end{equation}
The primary measure $U_1$, whose subscript denotes the one-chunk readout horizon rather than an evaluator world, evaluates a single executed policy chunk using raw signed progress toward the congruent target. We additionally report absolute regret
$L=1-\mathbb{E}_{p,r}\tfrac12[\tilde V_0(\mathbf{A}_{p0r})+\tilde V_1(\mathbf{A}_{p1r})]$, where $\tilde V_w$ is the registered monotone map of $V_w$ onto the bounded scale $[0,1]$.
Thus, $L$ measures shortfall from the normalized reference value $1$, not regret relative to the best achievable action. Unlike choice, $U_1$ and $L$ retain graded physical progress within a branch; unlike final task success, they evaluate the audited action before subsequent replanning can alter the outcome.

We therefore report direction, seed, and bidirectional results separately. A pair is \emph{fully reliable} only if both memory directions select their warranted branches under every registered action seed.

\subsection{Pair-Level Inference and Controls}
\label{subsec:inference}

\subsubsection{Statistical units}
The independent unit of a constructed audit is the \emph{pair cluster}; memories, common-seed queries, evaluator worlds, and replays are repeated observations within the same cluster. The natural-rollout test resamples ordinary episodes within layout strata. Models, tasks, and local-audit cohorts are never pooled into a single effect. Unavailable values are reported as missing rather than zeros.

\subsubsection{Irrelevant-history control}

\looseness=-1
We extend each pair to a two-by-two design over Semantic $\times$ Nuisance histories. Each semantic origin contributes two detour routes, $a$ and $b$, with different valid motions, equal native update counts, and the same restored final current. A semantic memory effect must satisfy the preregistered positive rule on both routes. Nuisance effects on choice and value, together with the Semantic $\times$ Nuisance interaction, are assessed by equivalence: the entire 90\% pair-bootstrap interval must lie within its preregistered margin. Merely failing to reject a difference is therefore insufficient.

\subsection{Native-Rollout Memory Interventions}
\label{sec:method-extensions}

The same separation principle extends to states reached during native rollouts through two intervention classes. A \emph{native-event history intervention} enrolls an ordinary rollout when a frozen, outcome-blind event rule fires. It fixes the enrolled \emph{recipient}'s current input and non-memory state, then replaces only the declared history carrier: the explicit bank, counters, and stream for Mem-0, or a lawfully regenerated history tape for BridgeVLA++. Replacement content is drawn either from the recipient's own history or from a \emph{donor} episode. Measured splice residuals bound the interpretation to the declared carrier rather than to an uncontrolled change in memory content alone.

A second intervention, \emph{matched-prefix restoration}, clears the bank at a native query boundary, allows the policy to execute for a fixed fault period, and restores a declared component at the next query. The remaining intervention budget is shared across compared conditions. Its endpoint is official full-episode success, which is distinct from the one-chunk physical value $U_1$.

Each intervention class fixes its intervention point and readout in advance. For native-event interventions, the endpoint is \emph{commitment}: a sustained approach toward one target confirmed over the remaining continuation. A transient earlier segment toward another target does not overturn this endpoint and is reported separately from both the first sustained direction and official success. We keep two clocks separate: the \emph{intervention stage}, which specifies the treated boundary, and the \emph{readout window}, which may be a fixed window, native action chunk, or official endpoint.

Pre-intervention non-memory state is fixed exactly; afterward, the native rollout continues, allowing memory-dependent computation to diverge naturally. Cleared-bank and no-restore conditions are therefore not current-only baselines: native updates continue after intervention, so a cleared bank refills from post-clear observations, including a fresh anchor for Mem-0. The restoration population is a preassigned stage-and-delay mixture. An episode whose assigned event is never reached retains its shared unperturbed outcome.

\section{Experiments}
\label{sec:experiments}
\looseness=-1
We evaluate CMA through four research questions designed to test progressively stronger claims about memory use:


\begin{itemize}
    \item \textbf{RQ1:} Are task success and action sensitivity sufficient evidence that memory guides an individual decision? (Sec.~\ref{sec:dissociation})
    \item \textbf{RQ2:} Does memory content exert closed-loop influence beyond the constructed audit, through history replacement and memory restoration? (Sec.~\ref{sec:native})
    \item \textbf{RQ3:} Does CMA behave as intended under known-answer calibration and irrelevant-history controls, and does the audited decision structure occur in untreated natural rollouts?  (Sec.~\ref{sec:controls})
\item \textbf{RQ4:} Does the distinction between action change and memory-guided choice persist on physical hardware? (Sec.~\ref{sec:physical})
\end{itemize}




\subsection{Experiment Setup}

\looseness=-1
\textit{Subjects.}
We audit the official Mem-0 \texttt{m1\_mix}~\cite{chen2026rmbench} on Put Back, Rearrange, and Swap, and the official BridgeVLA++~\cite{li2026bridgevlapp} on Put Back. The two policies are analyzed separately and never pooled. All simulated evaluations use frozen checkpoints, with no training, fine-tuning, or weight updates. For the physical deployment (Sec.~\ref{sec:physical}), task-specific Mem-0 execution modules are retrained for the dual-arm PiPER embodiment and frozen before auditing. This is embodiment adaptation rather than zero-shot transfer of the simulated checkpoint.

\textit{Qualification and cohorts.}
Both policies first pass a native qualification gate before entering any audit cohort: Mem-0 achieves $20/20$, $18/20$, and $19/20$ on Put Back, Rearrange, and Swap, respectively, while BridgeVLA++ passes its frozen 20-episode qualification gate. These episodes are excluded from all subsequent analyses. Native-event and restoration experiments use fresh, disjoint populations enrolled by frozen, outcome-blind rules; episodes that never reach the assigned enrollment event are retained. Table~\ref{tab:cohorts} summarizes the cohorts and independent units. All confirmatory intervals use 10,000 preregistered bootstrap draws over the reported denominators. Record-level cohort inputs and an anonymous recomputation kit are provided as supplementary artifacts.

\looseness=-1
\textit{Primary audited decision.}
The main constructed audit uses the later-Swap decision. In Swap, two visually identical blocks are moved among three trays through sequential transfers followed by a final button press. We place the audit after the common grasp-and-lift phase and before empty-tray selection, when the precommitment motion is complete and the branch oracle evaluates the decision. Two task-consistent histories therefore produce the same verified current input while warranting different branches. Official task success requires the complete swapped configuration; an individual transfer or transiently swapped geometry is not treated as task success.

\begin{table}[!htbp]
\caption{\textbf{Selected simulation cohorts and independent units.} Directions, histories, conditions, and seed repeats are within-unit observations; cohorts are never pooled. Units are pair clusters unless marked as episodes. M0: Mem-0; B++: BridgeVLA++ (deterministic queries); PB: Put Back. Calibration and natural-rollout cohorts: Sec.~\ref{sec:controls}; physical cohorts: Table~\ref{tab:physical}.}
\label{tab:cohorts}

\centering
\footnotesize
\setlength{\tabcolsep}{3pt}

\begin{tabular}{@{}lll@{}}
\toprule
Cohort & Units & Within each unit \\
\midrule

\multicolumn{3}{@{}l@{}}{
    \textit{RQ1: Constructed decision audits}
} \\
M0, PB
    & 64 pairs
    & 2 dir.$\times$3 seeds \\
M0, later Swap
    & 24 pairs
    & 2 dir.$\times$2 seeds \\

\addlinespace[3pt]
\multicolumn{3}{@{}l@{}}{
    \textit{RQ2: Native-rollout interventions}
} \\
Native event (B++, PB)
    & 24 pairs
    & 2 dir.$\times$1 query \\
Native event (M0, PB)
    & 12 pairs
    & 2 dir.$\times$2 seeds \\
Stage cohort (M0, PB)
    & \shortstack[l]{24 pairs\\(12/stage)}
    & 2 directions \\
Staged removal (M0, Swap)
    & 24 episodes
    & removal + restore \\
Anchor recovery (M0, Swap)
    & 36 episodes
    & restore + nested seeds \\

\addlinespace[3pt]
\multicolumn{3}{@{}l@{}}{
    \textit{RQ3: Irrelevant-history control}
} \\
Nuisance $2{\times}2$ (M0, PB)
    & 48 pairs
    & 4 histories$\times$3 seeds \\

\bottomrule
\end{tabular}
\end{table}
\subsection{When Action Change Misleads (RQ1)}
\label{sec:dissociation}

\begin{figure*}[t]
  \centering
  \includegraphics[width=0.95\textwidth]{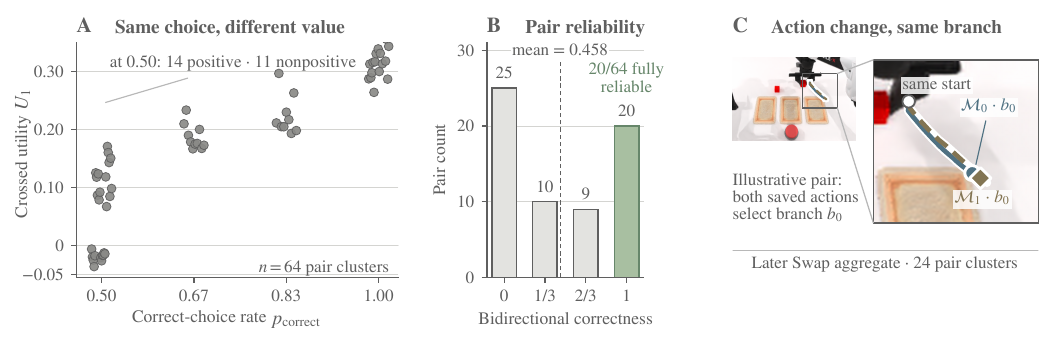}
  \caption{\textbf{Choice, physical value, and pair reliability describe the same saved
  actions differently.} (A) Put Back pairs at the same correct-choice rate differ in
  crossed utility; horizontal jitter is for visibility only. (B) Per-pair bidirectional
  correctness across the three registered action seeds, over the same 64 pair clusters.
  (C) Two saved 31-point paths from one later-Swap pair (magnified from the marked
  region) illustrate action change without branch change; cohort statistics come from
  the 24 pairs, not this single pair.}

  \label{fig:incremental}
\end{figure*}

\looseness=-1
To answer RQ1, we first examine the two metrics most commonly used to assess memory use in robot policies: task success and action sensitivity. CMA separates these metrics from the stronger question of whether the changed action selects the branch warranted by the remembered history. On the same frozen checkpoint, the two metrics dissociate in opposite directions: Put Back is correct on average but unreliable pair by pair, whereas later Swap changes every action without changing its branch (Table~\ref{tab:dissociation}).

\looseness=-1
\textbf{Put Back Evaluation.} Mem-0 passes native qualification ($20/20$), and all 64 valid audited pairs exhibit an action change across the two memories. Yet action sensitivity does not imply reliable branch selection. The correct-choice rate is $0.729$ with signed faithfulness $F=0.529$, while only $20/64$ pairs are fully reliable across all three registered action seeds. Matched-world physical value is positive ($U_1=0.185$), but absolute regret is $0.442$, showing that even average physical benefit does not imply reliable memory-guided decisions (Fig.~\ref{fig:incremental}A,B).

\textbf{Later Swap Evaluation.} Later Swap provides the complementary case. Every one of the 24 paired audits changes its action, yielding $S_{\mathrm{exec}}=0.059$, yet the selected branch never changes (Fig.~\ref{fig:incremental}C): $p_{\mathrm{correct}}=0.5$, $F=0$, and no pair is fully reliable. The policy therefore responds to memory without using it to distinguish the warranted branch. Together, the two tasks demonstrate that \emph{task success} and \emph{action change} provide different and incomplete evidence: success can conceal local decision failures, while sensitivity can occur without semantic use.
The dissociation is stable across the analysis grid. For Put Back, the qualitative ordering is unchanged over all five audit horizons (5--30 commands) and choice margins from $0.025$ to $0.075$; later Swap remains branch-invariant at every tested margin. Decision phase also matters: early Rearrange and Swap actions are memory-sensitive but fully ambiguous, whereas after the common grasp-and-lift phase, ambiguity vanishes and branch invariance becomes identifiable.
Put Back failures also cluster spatially: the correct-choice rate at target~3 is $0.115$, compared with $0.865$--$0.979$ at targets~0--2, with the same pattern in the held-out population. Because target location is never intervened on, this identifies a target-associated failure pattern rather than a geometric cause.

\begin{table}[t]
\caption{\textbf{Registered endpoints of the dissociation.} Constructed cohorts of Table~\ref{tab:cohorts}; preregistered 95\% pair-cluster bootstrap intervals. $S_{\mathrm{exec}}$: executed action divergence; $F$: signed faithfulness ($\neq 2p_{\mathrm{correct}}-1$ under nonzero ambiguity); regret: shortfall from the reference value $1$; fully reliable: correct in both directions under all seeds. Blank: structural values or endpoints without registered intervals.}

\label{tab:dissociation}
\centering
\footnotesize
\setlength{\tabcolsep}{5pt}
\begin{tabular}{@{}lll@{}}
\toprule
Endpoint & Put Back & later Swap \\
\midrule
$S_{\mathrm exec}$ & 0.054 \ci{0.051}{0.057} & 0.059 \ci{0.053}{0.065} \\
$p_{\mathrm correct}$ & 0.729 \ci{0.677}{0.781} & 0.5 \\
$F$ & 0.529 \ci{0.422}{0.633} & 0 \\
$U_1$ & 0.185 \ci{0.155}{0.215} & $-0.019$ \ci{-0.031}{-0.008} \\
Absolute regret & 0.442 & 0.772 \\
Fully reliable & 20/64 & 0/24 \\
\bottomrule
\end{tabular}
\end{table}

\begin{figure}[t!]
  \centering
  \includegraphics[width=0.9\columnwidth]{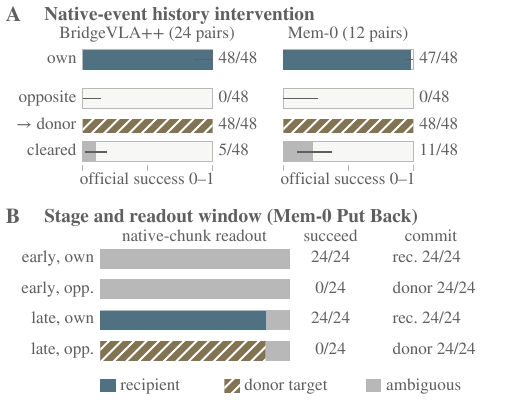}
  \caption{\textbf{Replacing the history bank at native events redirects behavior to the replaced content.} (A) Official success under own-history reconstruction, opposite-target history, and cleared banks, each policy analyzed separately; the hatched row is the share of opposite-history continuations committing to the donor target. Whiskers: 95\% cluster intervals; denominators follow Table~\ref{tab:cohorts}.
  (B) Native-chunk readout composition (12 fresh Mem-0 Put Back pairs per stage): every
  early readout is ambiguous, although commitment later follows the content; bank-cleared success is 5/24 early and 6/24 late.}

  \label{fig:native-event}
\end{figure}

\subsection{When Replacement Redirects and Restoration Recovers (RQ2)}
\label{sec:native}

RQ1 shows that local action change is insufficient evidence of memory use. We therefore test whether memory influences subsequent closed-loop behavior through native replacement and restoration.

\subsubsection{History replacement at native events}

At a frozen, outcome-blind event, we replace the enrolled recipient's declared history carrier with either its own history or a donor episode's opposite-target history, while keeping the current input and non-memory state fixed. Own-history reconstruction preserves official success in $48/48$ BridgeVLA++ and $47/48$ Mem-0 continuations; the single Mem-0 failure commits correctly but fails physically. In contrast, opposite-target history reduces official success to $0/48$ for each policy, while all 48 continuations per policy commit to the donor target (Fig.~4A). Thus, replacing memory with content encoding the opposite task history redirects subsequent behavior toward that content, even though it specifies the wrong target for the recipient episode. The policies remain analyzed separately, with pair clusters as the independent units (Table~1).

\textbf{Content-specificity controls.}
We next distinguish memory-content effects from the replacement operation itself. In a separate Mem-0 Put Back stage cohort, donor-corresponding first sustained directions are higher under opposite-target replacement than under bank clearing by $+0.583$ \ci{0.500}{0.708} at the early event and $+0.667$ \ci{0.542}{0.792} at the late event. Self-restore controls reproduce saved actions, query schedules, and tracked block trajectories exactly (Sec.~III-D). These controls support content-specific redirection rather than an effect attributable to perturbation alone.

\begin{figure}[tb]
  \centering
  \includegraphics[width=0.9\columnwidth]{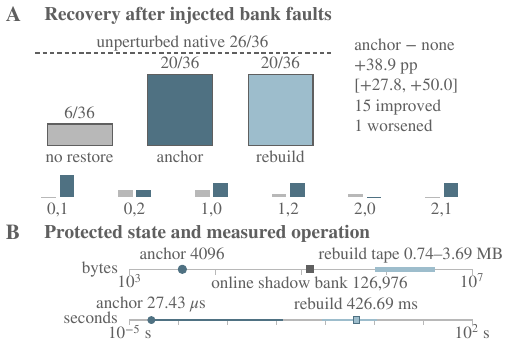}
  \caption{\textbf{Restoring a 4096-byte protected anchor recovers closed-loop success
  lost to injected bank faults.} (A) Official Swap success over the 36-episode assigned
  mixture after the injected whole-bank fault; the dashed line is the unperturbed native
  reference, and the strip shows the no-restore (gray) versus anchor (blue) contrast per
  ordered layout (six episodes each). (B) Protected state and measured restoration cost
  on log axes: 4096 bytes is the numeric tensor payload (serialized protection file:
  5321 bytes); timings cover the restoration operation only. Dots: medians; lines:
  observed min--max over the 32 primary recoveries. For storage reference, the online
  shadow bank holds 126{,}976 bytes.}

  \label{fig:restoration}
\end{figure}

\textbf{Readout Timing.}
At the frozen early event, all tested readout windows remain ambiguous, yet all 24 continuations under each history condition later commit to its encoded target. At the later center-grasp event, $21/24$ readouts per condition are already content-corresponding (Fig.~4B). Thus, an ambiguous instantaneous readout need not imply memory non-use when commitment occurs later. This also explains the difference from the constructed exact-current state, where BridgeVLA++ is history-sensitive but all 96 outcomes remain ambiguous ($U_1=-0.0026$ \ci{-0.0044}{-0.0010}): the two settings probe local decision dependence and subsequent closed-loop commitment, respectively.



\begin{figure*}[!t]
  \centering
  \includegraphics[width=0.95\textwidth]{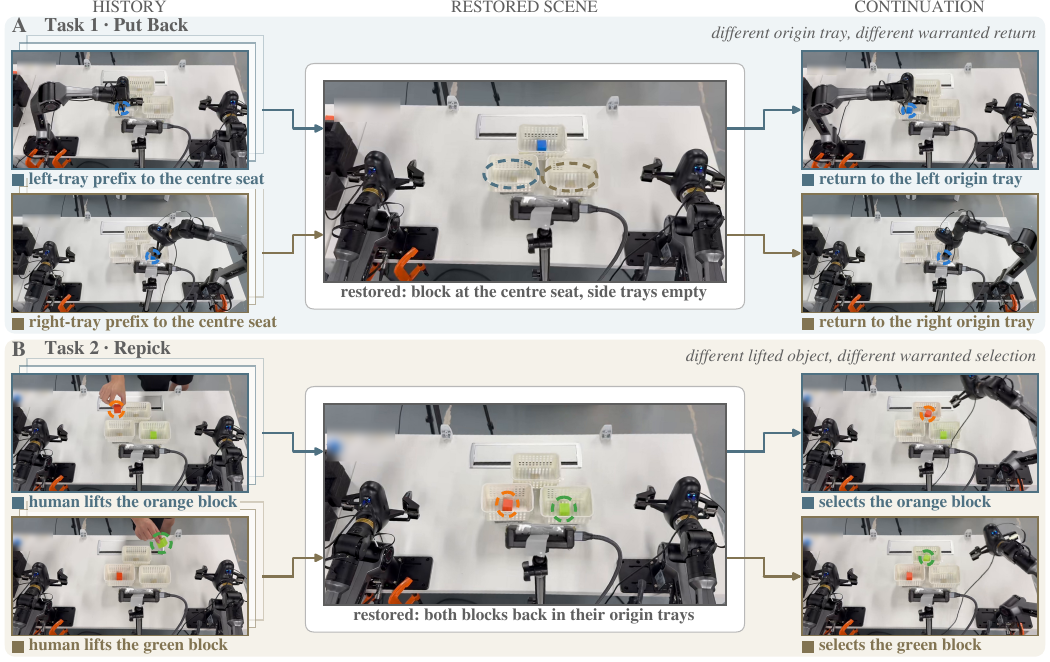}
    \caption{\textbf{The dual-arm physical audit: storyboards of both tasks.} Frames
  from two recorded demonstration runs per task (illustrative; all counts and intervals
  are in Table~\ref{tab:physical}). Colours mark the two histories, stacked outlines
  stored history, and dashed rings the decision-relevant objects and trays. (A) Put
  Back: the restored scene hides the origin, so the warranted return differs with the
  history. (B) Repick: the warranted selection differs with the lifted object. Equipment
  asset tags are blurred for double-blind review.}

  \label{fig:physical-tasks}
\end{figure*}

\subsubsection{Memory Restoration Recovers Success}

We then test whether restoring memory can recover behavior after an explicit memory fault. In 24 fresh Mem-0 Swap episodes, clearing the declared bank at a native query boundary reduces success at all three tested stages. Rebuilding the bank from the full observed history at the next query recovers all 12 native successes lost under removal, with no losses in the primary comparison.

\textbf{Anchor-only restoration.}
We next ask whether the smallest protected component tested can provide the same recovery. The primary confirmation uses an independent 36-episode cohort comparing no restoration, restoration of the original 4096-byte anchor tensor, and full rebuild from the same disturbed prefix. Intervention stages are crossed with 30- and 90-command fault delays; 32 of the 36 assigned episodes reach the fixed recovery boundary.
Official success is $6/36$ without restoration, $20/36$ with anchor restoration, and $20/36$ with full rebuild, compared with an unperturbed native reference of $26/36$ (Fig.~\ref{fig:restoration}A). Anchor restoration improves success over no restoration by $38.9$ percentage points (15 episodes improved, one worsened; layout-stratified 95\% bootstrap \ci{+27.8}{+50.0}). It matches full rebuild in this cohort, although neither restoration condition fully recovers the native success rate.

\looseness=-1
\textbf{Execution equivalence and cost.}
Mem-0 stores the anchor alongside a 30-entry recent-observation window. After a bank reset, native updates install a new anchor from post-fault observations. By the tested recovery boundary, the recent window has therefore been replenished from actual observations, so the disturbed and rebuilt banks differ only in the anchor tensor. Anchor restoration and full rebuild produce identical complete banks, states, actions, and subsequent trajectories in the tested comparisons. RMBench's ablation showed that anchor fusion matters for Swap~\cite{chen2026rmbench}. Here we test delayed restoration of the smallest protected component examined and measure its storage and restoration cost (Fig.~\ref{fig:restoration}B).

\subsection{Calibration and Controls (RQ3)}
\label{sec:controls}

We validate CMA against known semantic regimes, irrelevant history, and untreated behavior.
\subsubsection{Known-Answer Calibration}
Four known-answer calibrator cells, never pooled with Mem-0, return four distinguishable regimes. In a controlled update/interference task, transparent recurrent and retrieval policies show natural semantic-negative prevalences of 0.217 \ci{0.127}{0.321} and 0.100
\ci{0.061}{0.143} while crossed averages remain positive, so correct proxies coexist with individual semantic failures. Ordered-Route certifies the recurrent family ($S=0.441$, $F=1$, $U_1=0.544$; retrieval excluded with 2/5 seeds qualifying); on the native Put Back evaluator, a branch-aware local controller chosen on eight engineering pairs gives $F=1$ \ci{1}{1} with bidirectional correctness one over 48 pairs; and reusing one genuine action across balanced pseudo-memory labels returns the algebraic $F=U_1=0$, a consistency check rather than a general false-positive control.

\subsubsection{Irrelevant-History Control}
The irrelevant-history confirmation uses 48 pair clusters with three common seeds. Both mirror-detour routes pass the CMA positive rule, with  $p_{\mathrm correct}=0.753/0.757$. All three 90\% nuisance intervals---correct choice, matched-world value, and the Semantic$\times$Nuisance interaction---lie inside their preregistered margins (choice $\pm0.10$, value and interaction $\pm0.05$). The equivalence claim is limited to these two mirror-detour routes.

\subsubsection{Natural Occurrence Without Intervention}

Finally, we test whether the decision structure audited by CMA occurs in untreated behavior. An outcome-blind detector identifies the audited decision type in 86/96 natural episodes, labeling 52 as correct, 22 as ambiguous, and 12 as violating. The primary remaining-step contrast is unresolved  ($-870$
\ci{-4562}{2722}, layout-stratified episode bootstrap); the secondary final-success risk difference is $-0.550$ \ci{-0.720}{-0.375} and is treated as an observational association rather than a causal mechanism. The 24 constructed later-Swap pairs could not be exactly reconstructed in natural rollouts, so constructed and natural cohorts are never pooled.

\subsection{The Crossing Runs on Physical Hardware (RQ4)}
\label{sec:physical}

\looseness=-1
We reproduce the crossing protocol on a dual-arm PiPER platform (Fig.~\ref{fig:physical-tasks}) using two task-specific Mem-0 modules~\cite{chen2026rmbench}, retrained for the embodiment and frozen before evaluation. Put Back derives the warranted return from the robot's own manipulation history, whereas Repick derives the warranted selection from a completed external event. For each task, eight native history pairs share the current packet, non-memory state, and two action seeds; saved prefixes are replayed before closing the loop on fresh observations. Controlled resets are applied before each execution.


\begin{table}[t]
\caption{\textbf{Physical audit on the dual-arm PiPER cell: core readouts over eight
pairs / 32 executions per task.} Counts use operator-provided labels, not independent
blinded annotation; one further execution per task is a mixed/other case. Direction is
an auxiliary movement readout. Independent units are the eight pair clusters per task;
physical repeats and action seeds are within-unit observations.}
\label{tab:physical}
\centering
\footnotesize
\setlength{\tabcolsep}{5pt}
\begin{tabular}{@{}lll@{}}
\toprule
Readout & Put Back & Repick \\
\midrule
Target correct & 6/32 & 8/32 \\
Wrong target & 6/32 & 6/32 \\
No expressed selection & 19/32 & 17/32 \\
Manipulation completed & 9/32 & 10/32 \\
\quad thereof at the wrong target & 5/9 & 5/10 \\
Full success & 4/32 & 5/32 \\
History-consistent direction & 16/32 & 18/32 \\
Pairs correct in all four exec. & 0/8 & 0/8 \\
\bottomrule
\end{tabular}
\end{table}

\looseness=-1

Table~\ref{tab:physical} shows the same decision-level dissociation observed offline: only $6/32$ Put Back and $8/32$ Repick executions select the history-consistent target, while full success is $4/32$ and $5/32$, respectively; no pair is correct in all four executions. History-consistent direction occurs in $16/32$ and $18/32$ executions, showing that movement, selection, and task success can diverge. In matched-seed Put Back swaps, only $3/16$ direction changes follow the swapped origin, while $9/16$ show no change and $3/16$ move oppositely.


The hardware study also reveals a practical reproducibility gap. Offline recomputation exactly reproduces saved actions, whereas physical repeats with identical prefixes do not reproduce their direction categories. Moreover, action-seed variation ($0.0674$ rad joint RMS) is substantially larger than the memory-swap difference ($0.0076$ rad). Thus, the crossing protocol is reproducible at the policy-query level but remains subject to substantial closed-loop physical variability.


\section{Discussion and Conclusion}
\label{sec:discussion}

When does memory guide robot action? Our audit shows that neither task success nor action change provides a sufficient answer. A policy can change every action without changing its task branch, or select the correct branch on average while failing pair by pair, across directions, or across action seeds. The same dissociation appears on hardware under measured resets. Yet memory can exert genuine closed-loop influence: native-event bank replacement redirects commitment toward the replaced content, while restoring a single protected 4096-byte component recovers 38.9 points of success lost to injected bank faults.

CMA therefore provides a decision-level complement to episode-scored memory evaluation. It makes memory use testable at a fixed present through sensitivity, warranted choice, matched-world value, and reliability, while explicitly declaring the intervention state, readout window, and independent sampling unit. Its conclusions remain bounded to the audited checkpoints and tasks; two held-out comparisons showed no exclusive predictive gain over strong simple baselines. Ultimately, a memory-guided robot must do more than remember or react: it must reliably use its past to choose the behavior that the present alone cannot determine.

\bibliographystyle{IEEEtran}
\bibliography{references}
\end{document}